\documentclass[conference]{IEEEtran}
\IEEEoverridecommandlockouts
\usepackage{cite}
\usepackage{amsmath,amssymb,amsfonts}
\usepackage{algorithmic}
\usepackage{graphicx}
\usepackage{textcomp}
\usepackage{xcolor}
\def\BibTeX{{\rm B\kern-.05em{\sc i\kern-.025em b}\kern-.08em
    T\kern-.1667em\lower.7ex\hbox{E}\kern-.125emX}}
\usepackage{tabularx}
\usepackage{booktabs}
\usepackage{float}
\usepackage[utf8]{inputenc}
\DeclareUnicodeCharacter{2265}{$\geq$}

\usepackage{placeins}  % Garante o posicionamento de figuras e tabelas
\usepackage{float}      % Controle adicional para posicionamento

\usepackage{float}  % Para controlar a posição exata das tabelas
\usepackage{graphicx}  % Para incluir gráficos e tabelas

\usepackage[font=small,skip=0pt]{caption}

\usepackage{booktabs}  % Para tabelas mais elegantes
\usepackage{multicol}  % Para colunas adicionais em tabelas
\usepackage{multirow}  % Para células mescladas em tabelas
\usepackage{placeins}  % Forçar a ordem de floats (ex.: \FloatBarrier)

\renewcommand{\thetable}{\arabic{table}}

\author{
{\rm Giuliano Lorenzoni}\\
glorenzo@uwaterloo.ca\\
University of Waterloo\\
Waterloo, Ontario, Canada
\and
{\rm Ivens Portugal}\\
iportugal@uwaterloo.ca\\
University of Waterloo \\
Waterloo, Ontario, Canada

\and
{\rm Paulo Alencar}\\
palencary@uwaterloo.ca\\
University of Waterloo \\
Waterloo, Ontario, Canada

\and
{\rm Donald Cowan}\\
dcowan@uwaterloo.ca\\
University of Waterloo \\
Waterloo, Ontario, Canada
}

\begin{document}
\title{Exploring Variability in Fine-Tuned Models for Text Classification with DistilBERT} 
\maketitle

\begin{abstract}

This study presents an extensive evaluation of fine-tuning strategies for text classification using the DistilBERT model, specifically focusing on the distilbert-base-uncased-finetuned-sst-2-english variant. Through a structured experimental design, we examine the influence of hyperparameters—namely learning rate, batch size, and number of epochs—on key performance metrics, including accuracy, F1-score, and loss. Utilizing polynomial regression analyses in both absolute and relative frameworks, our approach captures foundational and incremental impacts of these hyperparameters, with a particular focus on fine-tuning adjustments relative to a baseline model.

Results highlight notable variability in metric outcomes based on hyperparameter configurations, illustrating that optimizing for one metric often leads to trade-offs impacting others. For example, while an increase in learning rate was associated with reduced loss in relative analysis (p = 0.027), it posed challenges to maintaining consistent accuracy improvements. Conversely, batch size demonstrated consistent significance for accuracy and F1-score in absolute regression (p = 0.028 and p = 0.005, respectively), yet had a limited impact on loss optimization (p = 0.170). Additionally, the interaction between epochs and batch size was especially critical for maximizing F1-score (p = 0.001), emphasizing the importance of hyperparameter interplay.

This interdependence underscores the need for fine-tuning strategies that address non-linear interactions among hyperparameters to achieve balanced performance across metrics. Our findings suggest that such variability and metric trade-offs are relevant considerations across subtasks and applications beyond text classification, extending to other areas in NLP and computer vision. This analysis not only informs fine-tuning strategies tailored to large language models but also highlights the importance of adaptive designs for broader model applicability.

\end{abstract}

\begin{IEEEkeywords}
Large language models, DistilBERT, fine-tuned models, text classification, variation.
\end{IEEEkeywords}

\section{Introduction}

The practice of fine-tuning has become a fundamental step in adapting \textit{Large Language Models} (LLMs) to effectively perform across a wide range of tasks. For these models, the ability to adjust hyperparameters to optimize performance in specific tasks not only improves accuracy and relevance but also tailors the model to the distinct needs of various contexts, such as text classification, natural language generation, and other applications in \textit{Natural Language Processing} (NLP) and \textit{Computer Vision} (CV). However, the diversity of tasks and subtasks, along with the variety of available models and hyperparameters, necessitates the development of fine-tuning strategies that are adapted to the particularities of each task and, most importantly, the specific model being utilized.

In this study, we seek to obtain practical insights for designing fine-tuning strategies tailored to the \textit{DistilBERT} model, exploring the impacts of essential hyperparameters—learning rate, batch size, and number of epochs—on performance metrics, including accuracy, F1-score, and loss. This analysis considers not only the isolated behaviors of these relationships in fine-tuned models but also performance differences relative to the baseline model, allowing a detailed view of both foundational and incremental effects. This approach facilitates the identification of trade-offs and interdependencies among metrics, which is crucial for formulating balanced fine-tuning strategies.

To achieve these objectives, we structure the paper as follows: in the \textit{Related Work} section, we contextualize fine-tuning practices and limitations in LLMs and discuss recent studies on the impact of hyperparameters on different metrics. In \textit{Experimental Design}, we detail the methodology used, including data collection and the process of selecting hyperparameters and evaluation metrics. The \textit{Results} section presents the main findings of the analysis, focusing on the relationship between hyperparameters and metrics. Finally, in \textit{Discussion} and \textit{Conclusions}, we discuss the key challenges identified and suggest future directions that could benefit a variety of models and tasks in NLP and beyond.

% Lorem ipsum dolor sit amet, consectetur adipiscing elit. Nulla vel eros vitae eros feugiat lacinia. Cras vehicula volutpat ipsum, id tincidunt nisi porttitor ac. Nam sagittis enim in turpis sollicitudin, ut pellentesque magna vehicula. Donec et turpis vestibulum, blandit lorem et, egestas massa. Duis ullamcorper, augue vel consectetur feugiat, ipsum risus eleifend orci, a dignissim nisi odio ut nisl. Sed dictum sapien ac lorem vulputate, nec suscipit nisl iaculis. Nulla facilisi. Suspendisse potenti.

% Phasellus ut lacus et orci malesuada vehicula. Maecenas feugiat metus sed magna tincidunt, non vulputate dolor tincidunt. Ut scelerisque facilisis nibh ut suscipit. Nulla facilisi. Vestibulum vitae fringilla felis, at hendrerit erat. Praesent non odio eget lacus consequat tincidunt. Aenean sollicitudin malesuada magna a maximus. Nam eget tristique tortor. In faucibus, urna ut euismod laoreet, magna magna finibus nisi, sit amet luctus ligula ante a risus.

% Vivamus sagittis erat id vestibulum dapibus. Phasellus in mi non lectus euismod gravida. Integer aliquet, erat sed tempor iaculis, magna orci mollis mi, at tempor turpis risus id lorem. In interdum, turpis quis ultricies blandit, ex ligula pretium metus, et fermentum lacus ligula ut velit. Donec tincidunt augue at ex efficitur ultricies. Donec at diam et sem suscipit bibendum. Nunc consectetur tellus ligula, et scelerisque ligula laoreet sit amet. Sed sit amet scelerisque lorem. Nulla pharetra nisi vitae lacus laoreet, eget pulvinar sem vehicula. Integer ac dolor ut ex pharetra venenatis ut ac ipsum.

\section{Related Work} \label{sec:related}

The fine-tuning of Large Language Models (LLMs) for specific applications has become a prominent focus in recent research, with studies examining a variety of strategies, hyperparameters, and domains. Here, we discuss several key contributions and compare them with our study, highlighting how our work uniquely explores multiple hyperparameters and performance metrics to derive actionable fine-tuning insights, particularly for the DistilBERT model in text classification.

\subsection{Learning Rate Tuning in LLMs}

The work on \textit{Rethinking Learning Rate Tuning in the Era of Large Language Models} \cite{jin2023rethinking} focuses on the critical role of learning rate in fine-tuning. The study proposes LRBench++, a tool for evaluating learning rate policies across LLMs and traditional DNNs, highlighting the need for tailored learning rate policies for LLMs. In contrast, our study expands beyond a singular focus on learning rate to evaluate the combined effects of learning rate, batch size, and epochs. Additionally, our approach incorporates both absolute and relative analyses of accuracy, F1-score, and loss to provide a holistic view of hyperparameter impacts on DistilBERT’s performance in text classification.

\subsection{Fine-Tuning Strategies for Domain-Specific Tasks}

A recent study, \textit{Evaluating the Effectiveness of Fine-Tuning Large Language Models for Domain-Specific Tasks} \cite{dabhi2024evaluating}, fine-tuned LLaMA 2 using two distinct methods on a migration-related news dataset, illustrating the impact of self-supervised versus direct Q\&A fine-tuning strategies. While this study concentrates on strategy effectiveness in aligning the model with domain-specific content, our work evaluates a broader range of hyperparameters and their trade-offs across core metrics. Unlike their focus on dataset alignment, we derive generalized fine-tuning strategies through polynomial regression analyses, applicable across different datasets and tasks.

\subsection{Parameter-Efficient Techniques for Domain Optimization}

\textit{Fine-Tuning Large Language Models for Task-Specific Data} \cite{zheng2024fine}examines the use of Parameter-Efficient Fine-Tuning (PEFT) and Quantized Low-Rank Adaptation (QLoRA) techniques to fine-tune LLaMA 2 and Falcon 7B models for an e-commerce dataset. This study demonstrated the contextual accuracy benefits of customized fine-tuning. In contrast, our study targets the foundational impacts of basic hyperparameters (learning rate, batch size, and epochs) on multiple performance metrics, producing insights for balanced tuning strategies rather than domain-specific optimizations.

\subsection{Supporting Fine-Tuning Through Automated Systems}

The study \textit{Fine-Tune it Like I'm Five: Supporting Medical Domain Experts in Training NER Models Using Cloud, LLM, and Auto Fine-Tuning} \cite{hartmann2023fine}introduces CRM4NER, a cloud-based system for managing, training, and fine-tuning Named Entity Recognition (NER) models. It incorporates automatic hyperparameter tuning and context-aware fine-tuning recommendations generated via a Large Language Model (LLM), aiming to support domain experts who may lack expertise in Machine Learning (ML). While CRM4NER emphasizes a user-friendly ML management system, our study directly assesses the fine-tuning effects of hyperparameters on model metrics, guiding users on tuning DistilBERT for optimal performance without requiring complex management systems.

\subsection{Fine-Tuning for Low-Resource Languages}

\textit{Fine Tuning LLMs for Low Resource Languages} \cite{joshi2024fine} addresses fine-tuning strategies for optimizing LLMs in low-resource language settings, exploring language-specific techniques such as LoRA, QLoRA, instruction tuning, and Representation Fine-Tuning (ReFT). This study emphasizes language-specific tuning needs, contrasting with our approach, which examines hyperparameter impacts on DistilBERT’s text classification performance regardless of language resources. We focus on understanding the metric-specific influences of core hyperparameters, providing insights that are applicable across diverse linguistic contexts.

\subsection{Optimizing LLM Performance at Scale}

\textit{Achieving Peak Performance for Large Language Models: A Systematic Review} \cite{rostam2024achieving} provides a comprehensive systematic literature review (SLR) of optimization and acceleration methods for large language models (LLMs), focusing on balancing high performance with practical constraints. While this review categorizes optimization strategies into LLM training, inference, and system serving, our work offers a more focused analysis of hyperparameters that directly affect DistilBERT’s classification performance. Our study contributes a metric-specific understanding of hyperparameter trade-offs, complementing high-level efficiency optimizations discussed in the review.

\subsection{Detecting Bots with Transformer-Based Models}

The study \textit{Fine-Tuned Understanding: Enhancing Social Bot Detection With Transformer-Based Classification} \cite{sallah2024fine} uses transformer-based models like BERT and GPT-3 to detect social media bots, achieving high F1-scores through fine-tuning. Unlike this bot-detection focus, our research evaluates the interplay of hyperparameters on general performance metrics. Our work seeks to balance multiple metrics rather than targeting domain-specific detection accuracy, expanding the application of our insights beyond niche uses.

\subsection{LLM Techniques for Document Understanding}

\textit{LayoutLLM: Layout Instruction Tuning with Large Language Models for Document Understanding} \cite{luo2024layoutllm} uses a layout-aware fine-tuning approach for document understanding, incorporating layout-aware pre-training and supervised fine-tuning to enhance the comprehension of document structures. Our study differs by exploring fundamental hyperparameter adjustments rather than layout-specific tuning strategies, providing adaptable insights for general text classification applications across various NLP tasks.

\subsection{Fine-Tuning DistilBERT in Classification Tasks}

Several studies have investigated fine-tuning DistilBERT for specific applications:
\begin{itemize}
    \item \textbf{Cyberbullying Detection in Social Networks: A Comparison Between Machine Learning and Transfer Learning Approaches} \cite{teng2023cyberbullying} used DistilBERT embeddings to detect abusive content with high F1-scores.
    \item \textbf{Hate Speech and Target Community Detection in Nastaliq Urdu Using Transfer Learning Techniques} \cite{malik2024hate} applied DistilBERT to identify hate speech in low-resource languages, demonstrating the model’s adaptability in multilingual contexts.
    \item \textbf{Depression Classification From Tweets Using Small Deep Transfer Learning Language Models} \cite{rizwan2022depression} employed DistilBERT for social media-based depression classification, comparing it with smaller models.
    \item \textbf{AGI-P: A Gender Identification Framework for Authorship Analysis Using Customized Fine-Tuning of Multilingual Language Model} \cite{sarwar2024agi} developed AGI-P, fine-tuning DistilBERT for gender identification in a multilingual setting.
    \item \textbf{Performance Analysis of Federated Learning Algorithms for Multilingual Protest News Detection Using Pre-Trained DistilBERT and BERT} \cite{riedel2023performance} fine-tuned DistilBERT within a Federated Learning framework for protest news detection.
\end{itemize}

These studies leverage DistilBERT for specific text classification applications but focus primarily on single-task or dataset-specific fine-tuning effects. Our study, however, explores hyperparameter-driven variability across multiple metrics (accuracy, F1-score, and loss) in DistilBERT’s performance. By examining both absolute and incremental changes from a baseline model, our research uniquely contributes a comprehensive evaluation framework that informs adaptable fine-tuning strategies for balanced performance.

% -----xxxxxxx---------

\section{Experiment Design}
This section details the methodology employed in selecting and evaluating models for text classification, including data collection, model selection, hyperparameters, evaluation metrics, and statistical methods.

\subsection{Data Collection}
The data collection process involved identifying and extracting information about hyperparameters and performance metrics from models available on the Hugging Face platform. Our goal was to select NLP models specialized in text classification and gather relevant data from associated files. 

\begin{itemize}
    \item \textbf{Model Filtering:} Selected the Natural Language Processing (NLP) category from Hugging Face’s available options.
    \item \textbf{Task Selection:} Filtered for Text Classification, yielding 30 available models.
    \item \textbf{Base Model Selection:} Chose DistilBERT (distilbert-base-uncased-finetuned-sst-2-english)  due to its popularity and high number of downloads.
    \item \textbf{Fine-Tuned Models Extraction:} 
    \begin{itemize}
        \item Retrieved detailed information from 55 fine-tuned models associated with DistilBERT.
        \item \textbf{Data Sources:} Extracted from README.md and config.json files for each model.
    \end{itemize}
\end{itemize}

\subsection{Model Selection and Task Design}
The model selection and task design aimed to explore the efficiency and performance of fine-tuned models in text classification.

\begin{itemize}
    \item \textbf{Base Model Selection:} Chose DistilBERT (distilbert-base-uncased-finetuned-sst-2-english) for its efficiency and popularity.
    \item \textbf{Objective:} Evaluated the impact of different hyperparameter configurations and compared results with other fine-tuned models and NLP tasks.
    \item \textbf{Task Design:} 
    \begin{itemize}
        \item Evaluated all 55 fine-tuned models to ensure consistent analysis across metrics.
    \end{itemize}
\end{itemize}

\subsection{Hyperparameters and Evaluation Metrics}

We began our analysis with a comprehensive set of nine hyperparameters and nine evaluation metrics, following the initial configurations recommended in the model documentation. To streamline our focus, we refined this selection based on the frequency of observations across the 55 models analyzed. This approach enabled us to prioritize hyperparameters and metrics that consistently appeared and provided statistically significant insights into model performance.

\subsection{Selection of Evaluation Metrics and Hyperparameters for Analysis}
Due to the limited statistical significance observed across a broader range of metrics and hyperparameters, our study focused on metrics and hyperparameters with a sufficient number of observations to allow for statistically significant fine-tuning strategies. This selection aimed to ensure a robust analysis of the most impactful elements on model performance.

\begin{itemize}
    \item \textbf{Evaluation Metrics:} We concentrated on three key metrics that provided enough observations to support meaningful statistical exploration:
    \begin{itemize}
        \item \textbf{Accuracy} - Selected as a primary measure for the overall correctness of model predictions, offering insight into general model performance.
        \item \textbf{F1-Score} - Focused on due to its balanced consideration of precision and recall, particularly important for assessing performance in cases with class imbalance.
        \item \textbf{Loss} - Included as it quantifies model prediction error, essential for evaluating the model’s error minimization during training and validation.
    \end{itemize}

    \item \textbf{Hyperparameters:} Three core hyperparameters were chosen, as they provided enough data points to examine statistically significant trends in fine-tuning:
    \begin{itemize}
        \item \textbf{Learning Rate} - An influential parameter for training speed and stability, providing sufficient variability in observations for meaningful analysis.
        \item \textbf{Batch Size} - Affects memory usage and convergence, offering adequate data to explore its relationship with model generalization and performance.
        \item \textbf{Number of Epochs} - Key in controlling the number of training iterations over the dataset, providing essential insights into learning and convergence patterns.
    \end{itemize}
\end{itemize}

By focusing exclusively on these metrics and hyperparameters, we ensured a statistically valid basis for identifying consistent patterns and potential fine-tuning strategies. This approach provides a clearer understanding of core dependencies and allows for targeted adjustments in fine-tuning aimed at optimizing model performance in a statistically sound manner.

\subsection{Statistical Methods and Tools}
To assess the impact of selected hyperparameters on evaluation metrics, we employed polynomial regression analyses across two versions—Absolute and Relative. These analyses aimed to identify patterns, correlations, and potential optimizations through detailed regression models, heatmaps, scatter plots, and regression plots.

\begin{itemize}
    \item \textbf{Polynomial Regression Analyses:} We conducted polynomial regressions to examine the influence of hyperparameters (Learning Rate, Batch Size, and Number of Epochs) on key evaluation metrics (Accuracy, F1-Score, and Loss) in two distinct versions:
    \begin{itemize}
        \item \textbf{Absolute Regression:} This version utilized the absolute values of both the metrics and hyperparameters. It aimed to reveal the direct effects of hyperparameter values on model performance metrics (Accuracy, F1-Score, Loss) as observed post-training. Absolute regression thus reflects the foundational impact of hyperparameters on baseline metric performance.
        
        \item \textbf{Relative Regression:} In the relative regression, we used the differences in metric values between each fine-tuned model and the baseline model, as well as the differences in the hyperparameters applied in fine-tuned versus baseline models. This version provides insights into the incremental improvements and optimization effects of fine-tuning adjustments, specifically measuring how hyperparameter variations contribute to performance gains beyond the baseline.
    \end{itemize}

    \item \textbf{Heatmaps and Scatter Plots:} We generated heatmaps to visualize correlations between hyperparameters and metrics, identifying patterns that could inform fine-tuning adjustments. Scatter plots illustrated the spread and potential trade-offs between hyperparameters and metric outcomes, facilitating a more granular understanding of parameter interactions.

    \item \textbf{Regression Plots:} In conjunction with the polynomial regression analyses, we produced regression plots to visualize fitted regression lines and residuals. These plots provided additional insights into model fit and variability, helping to identify areas where hyperparameter adjustments yield statistically significant improvements.
\end{itemize}

\subsection{Consistency Evaluation between Metrics (Accuracy, F1, and Loss)}

The objective of this subsection is to systematically evaluate the consistency of optimization strategies across the key metrics of accuracy, F1, and loss. By analyzing both absolute and relative regression results, we aim to identify patterns that highlight common and divergent impacts of hyperparameters across these metrics. This approach allows us to observe how adjustments to batch size, learning rate, and epochs influence each metric and to determine whether improvements in one metric, such as accuracy, align or conflict with performance outcomes in F1 and loss. Ultimately, this evaluation provides insights into cross-metric coherence, informing the design of balanced fine-tuning strategies that prioritize comprehensive model performance.

\subsection{Mapping Fine-Tuning Strategies and Recommendations for Improvements}

This subsection synthesizes the insights captured for each metric based on both absolute and relative regression analyses to formulate actionable fine-tuning strategies. By mapping hyperparameter impacts on accuracy, F1, and loss, we can derive specific, data-driven recommendations that address the unique optimization needs of each metric while also accounting for their cross-effects. The objective is to enable a strategic approach to fine-tuning that emphasizes incremental gains in each metric while maintaining overall model stability. These strategies highlight which parameters to prioritize initially, such as batch size and epochs, and where to make final refinements, particularly with learning rate adjustments, to achieve balanced improvements in accuracy, F1, and loss.

\subsection{Computational Packages}
The implementation of our analyses leveraged several computational packages and libraries to streamline data processing, model training, statistical analysis, and visualization. Below is a summary of the key packages used in our study:

\begin{itemize}
    \item \textbf{Python Core Libraries:}
    \begin{itemize}
        \item \texttt{NumPy} - Provided efficient handling of numerical data and supported array-based computations essential for data manipulation and model input formatting.
        \item \texttt{Pandas} - Facilitated data manipulation, cleaning, and preparation, allowing us to manage datasets, perform feature engineering, and store results for analysis.
    \end{itemize}

    \item \textbf{Statistical Analysis and Metrics:}
    \begin{itemize}
        \item \texttt{SciPy} - Enabled statistical testing and analysis, including regression calculations and significance testing on hyperparameter impacts.
        \item \texttt{Scikit-Learn} - Supported computation of evaluation metrics (such as Accuracy, F1-score, Precision, and Recall) and provided tools for model evaluation, data splitting, and scaling.
    \end{itemize}
    
    \item \textbf{Visualization and Plotting:}
    \begin{itemize}
        \item \texttt{Matplotlib} and \texttt{Seaborn} - Used to generate detailed visualizations, including heatmaps, scatter plots, and regression plots. These visualizations facilitated a clearer understanding of hyperparameter effects and metric distributions across different configurations.
    \end{itemize}
\end{itemize}    

\section{Results}

% Lorem ipsum dolor sit amet, consectetur adipiscing elit. Nulla vel eros vitae eros feugiat lacinia. Cras vehicula volutpat ipsum, id tincidunt nisi porttitor ac. Nam sagittis enim in turpis sollicitudin, ut pellentesque magna vehicula. Donec et turpis vestibulum, blandit lorem et, egestas massa. Duis ullamcorper, augue vel consectetur feugiat, ipsum risus eleifend orci, a dignissim nisi odio ut nisl. Sed dictum sapien ac lorem vulputate, nec suscipit nisl iaculis. Nulla facilisi. Suspendisse potenti.

% Phasellus ut lacus et orci malesuada vehicula. Maecenas feugiat metus sed magna tincidunt, non vulputate dolor tincidunt. Ut scelerisque facilisis nibh ut suscipit. Nulla facilisi. Vestibulum vitae fringilla felis, at hendrerit erat. Praesent non odio eget lacus consequat tincidunt. Aenean sollicitudin malesuada magna a maximus. Nam eget tristique tortor. In faucibus, urna ut euismod laoreet, magna magna finibus nisi, sit amet luctus ligula ante a risus.

\subsection{Relationships Between Dependent and Independent Variables}

This section presents the results of analyzing the relationships between the dependent variables—accuracy, F1-score, and loss—and the independent variables (learning rate, batch size, and number of epochs). We report both absolute and relative regression analyses, along with heatmaps and scatter plots that help visualize the correlations and spreads across different hyperparameter configurations.

\subsubsection{Accuracy – Comparison of Absolute and Relative Regressions}

\paragraph{Regression Results}

\begin{itemize}
    \item \textbf{Absolute Regression}

    \begin{table}[H]
    \centering
    \scriptsize
    \renewcommand{\arraystretch}{1.1}
    \setlength{\tabcolsep}{4pt}
    \begin{tabular}{lrrrrr}
    \toprule
    Variable & Coef & Std Err & t & P>|t| & [0.025, 0.975] \\
    \midrule
    const & 0.1815 & 0.048 & 3.808 & 0.000 & [0.086, 0.277] \\
    x1 & -9595.1379 & 6952.389 & -1.380 & 0.178 & [-23800.33, 4624.095] \\
    x2 & 0.0773 & 0.007 & 10.601 & 0.000 & [0.062, 0.092] \\
    x3 & 0.0061 & 0.017 & 0.365 & 0.717 & [-0.028, 0.040] \\
    x4 & 3.705e+08 & 1.14e+08 & 3.238 & 0.003 & [1.36e+08, 6.05e+08] \\
    x5 & -0.0125 & 127.519 & -9.79e-05 & 1.000 & [-260.818, 260.793] \\
    x6 & -724.4980 & 588.683 & -1.231 & 0.228 & [-1928.490, 479.494] \\
    x7 & -0.0010 & 0.000 & -9.195 & 0.000 & [-0.001, -0.001] \\
    x8 & -0.0002 & 0.000 & -0.674 & 0.506 & [-0.001, 0.000] \\
    x9 & 0.0001 & 6.76e-05 & 1.729 & 0.095 & [-2.14e-05, 0.000] \\
    \bottomrule
    \end{tabular}
    \caption{Absolute Regression Results: Accuracy}
    \label{tab:regression_absolute_accuracy}
    \end{table}

    \begin{itemize}
        \item Batch size ($X_2$) showed a coefficient of $0.0773$ ($p = 0.000$), indicating a significant positive impact on accuracy.
        \item Num\_epochs ($X_3$) had a coefficient of $0.0061$ ($p = 0.717$), showing no statistical significance.
        \item Learning rate ($X_7$) exhibited a negative coefficient of $-0.001$ ($p = 0.000$), suggesting that higher learning rates can hinder performance.
    \end{itemize}

    \item \textbf{Relative Regression (Differences)}

    \begin{table}[H]
    \centering
    \scriptsize
    \renewcommand{\arraystretch}{1.1}
    \setlength{\tabcolsep}{4pt}
    \begin{tabular}{lrrrrr}
    \toprule
    Variable & Coef & Std Err & t & P>|t| & [0.025, 0.975] \\
    \midrule
    const & 0.7200 & 0.222 & 3.242 & 0.003 & [0.264, 1.177] \\
    x1 & -3995.6276 & 3471.177 & -1.151 & 0.260 & [-1.11e+04, 3139.479] \\
    x2 & 0.0230 & 0.008 & 2.996 & 0.006 & [0.007, 0.039] \\
    x3 & 0.0010 & 0.008 & 0.119 & 0.906 & [-0.016, 0.018] \\
    x4 & 4.605e+08 & 1.31e+08 & 3.521 & 0.002 & [1.92e+08, 7.29e+08] \\
    x5 & -88.3120 & 149.785 & -0.590 & 0.561 & [-396.199, 219.575] \\
    x6 & -1213.2849 & 723.274 & -1.677 & 0.105 & [-2699.997, 273.427] \\
    x7 & -0.0014 & 0.000 & -3.181 & 0.004 & [-0.002, -0.001] \\
    x8 & -0.0006 & 0.000 & -1.401 & 0.173 & [-0.001, 0.000] \\
    x9 & 0.0002 & 9.73e-05 & 2.016 & 0.054 & [-3.84e-06, 0.000] \\
    \bottomrule
    \end{tabular}
    \caption{Relative Regression Results: Accuracy}
    \label{tab:regression_relative_accuracy}
    \end{table}
    
    \begin{itemize}
        \item Learning rate ($X_7$) stood out with a coefficient of $-0.0014$ ($p = 0.004$), highlighting the critical role of fine-tuning for incremental improvements.
        \item Batch size ($X_2$) remained relevant, with a positive coefficient of $0.0230$ ($p = 0.006$).
    \end{itemize}
\end{itemize}

\paragraph{Visual Analysis and Insights}

The previous results are complemented by the Heatmaps (Figures 1 and 2) and Polynomial Regression Plots (Figures 2 and 4).

\begin{figure*}[h!]
\centering
\begin{minipage}{0.40\textwidth}
    \centering
    \includegraphics[width=\textwidth]{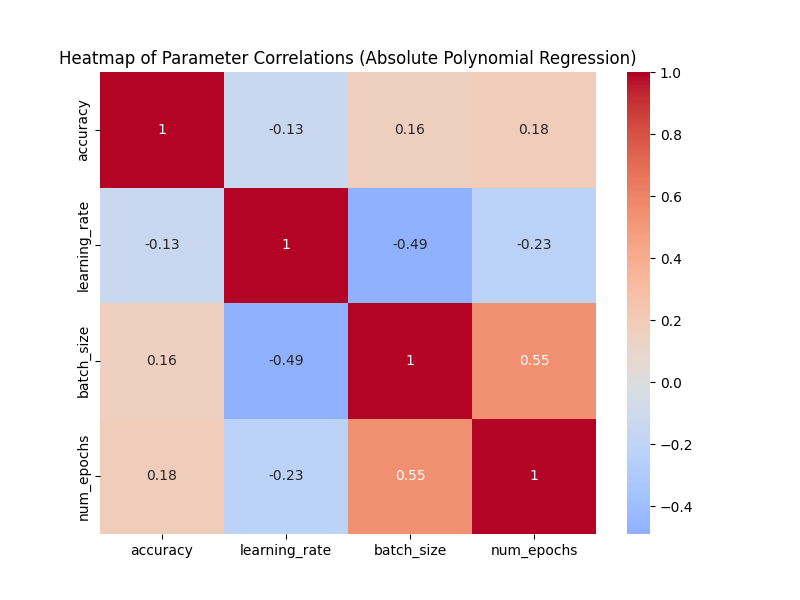}
    \caption{Heatmap of Parameter Correlations (Absolute Polynomial Regression)}
    \label{fig:heatmap_absolute}
\end{minipage}%
\hfill
%\end{figure*}
%\begin{figure*}[h!]
%\centering
\begin{minipage}{0.40\textwidth}
    \centering
    \includegraphics[width=\textwidth]{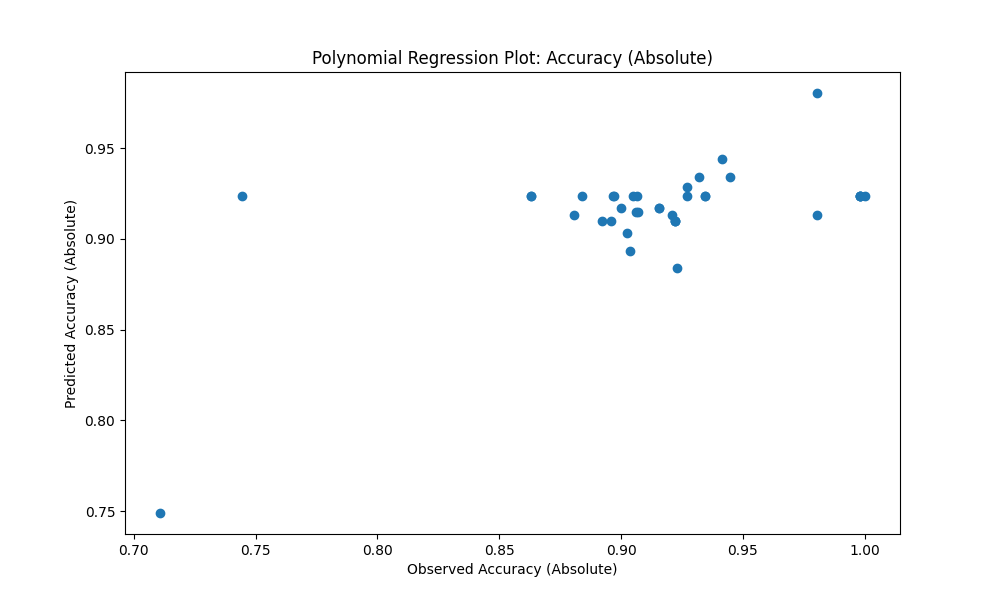}
    \caption{Polynomial Regression Plot: Accuracy (Absolute)}
    \label{fig:polynomial_regression_plot_absolute}
\end{minipage}
\end{figure*}

% \begin{figure*}[h!]
% \centering
% %\vspace{0.5cm}
% \begin{minipage}{0.40\textwidth}
%     \centering
%     \includegraphics[width=\textwidth]{figures/absolute_polynomial_scatter_plots_ALT.png}
%     \caption{Scatter Plots: Accuracy vs Learning Rate, Batch Size, and Number of Epochs}
%     \label{fig:scatter_plots_absolute}
% \end{minipage}%
% \hfill
% %\end{figure*}
% %\begin{figure*}[h!]
% %\centering
\begin{figure*}[h!]
\centering
\begin{minipage}{0.40\textwidth}
    \centering
    \includegraphics[width=\textwidth]{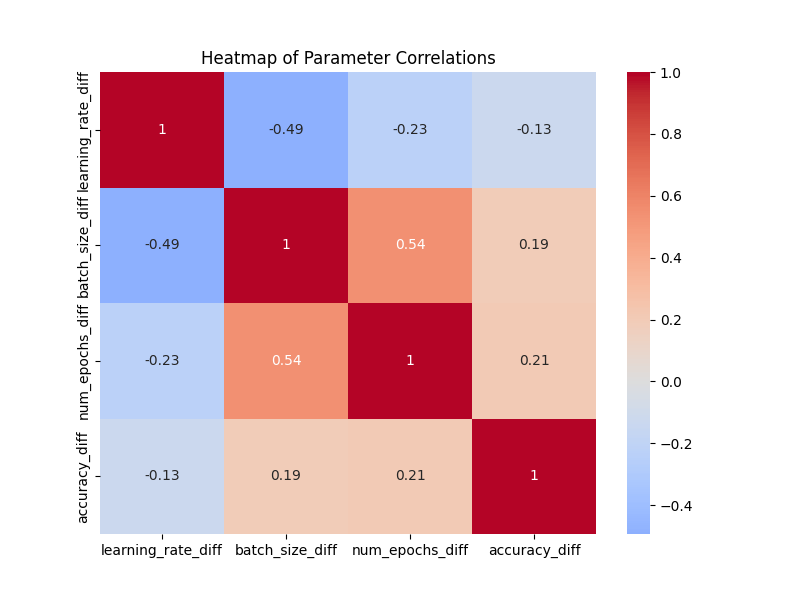}
    \caption{Heatmap of Parameter Correlations (Relative Polynomial Regression)}
    \label{fig:heatmap_relative}
\end{minipage}
\hfill
%\begin{figure*}[h!]
%\centering
\begin{minipage}{0.40\textwidth}
    \centering
    \includegraphics[width=\textwidth]{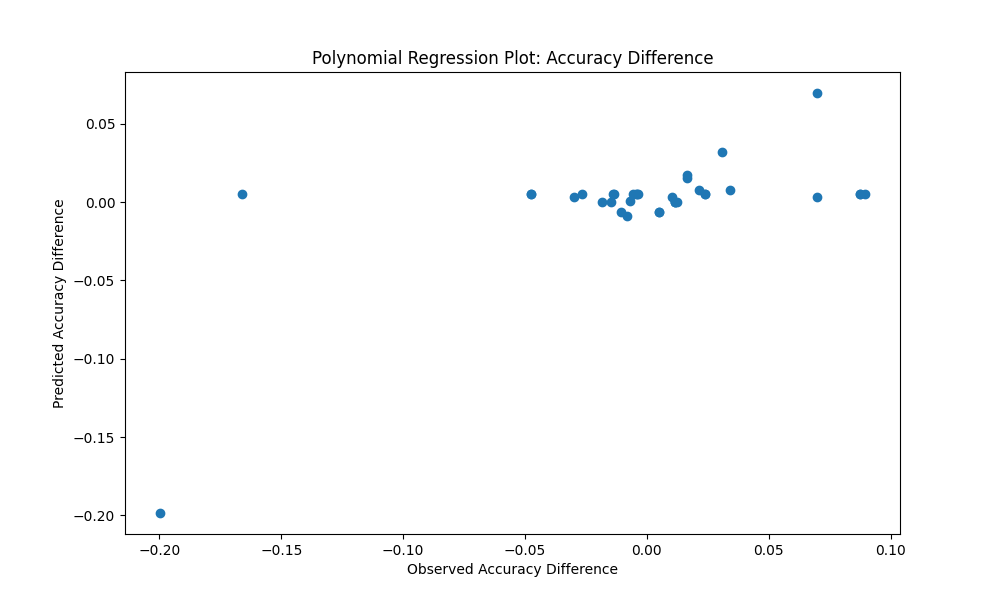}
    \caption{Polynomial Regression Plot: Accuracy Difference (Relative)}
    \label{fig:polynomial_regression_plot_relative}
\end{minipage}%
\end{figure*}

\paragraph{Consolidated Findings and Practical Insights on Accuracy}

The analysis reveals that the learning rate ($X_7$) significantly influences accuracy, with both absolute and relative analyses underscoring its critical role. A negative coefficient (-0.001, p = 0.000) in absolute terms highlights that higher learning rates can degrade accuracy, warranting cautious adjustments to prevent performance loss. In the relative analysis, fine-tuning learning rate adjustments produced notable incremental improvements (coefficient = -0.0014, p = 0.004), emphasizing its value in fine-tuning relative to baseline performance. Batch size ($X_2$), meanwhile, proved foundational for establishing accuracy during initial training, contributing to consistent performance. Combined, batch size and learning rate adjustments offer balanced accuracy gains, where initial selection should prioritize an optimal batch size to stabilize accuracy early on, followed by incremental learning rate increases. During fine-tuning, learning rate sensitivity demands precise, gradual adjustments for continued performance enhancement relative to the baseline. This structured, stage-wise approach aligns hyperparameter selection with both robust training and fine-tuning objectives.

\subsubsection{F1 Score – Comparison of Absolute and Relative Regressions}

\paragraph{Regression Results}

\begin{itemize}
    \item \textbf{Absolute Regression}

    \begin{table}[H]
    \centering
    \scriptsize
    \renewcommand{\arraystretch}{1.1}
    \setlength{\tabcolsep}{4pt}
    \begin{tabular}{lrrrrr}
    \toprule
    Variable & Coef & Std Err & t & P>|t| & [0.025, 0.975] \\
    \midrule
    x1 & 9934.1284 & 1.79e+04 & 0.556 & 0.587 & [-2.84e+04, 4.82e+04] \\
    x2 & 0.0452 & 0.018 & 2.459 & 0.028 & [0.006, 0.085] \\
    x3 & 0.1811 & 0.054 & 3.370 & 0.005 & [0.066, 0.296] \\
    x4 & 3.113e+08 & 2.53e+08 & 1.232 & 0.238 & [-2.31e+08, 8.53e+08] \\
    x5 & -276.8026 & 236.510 & -1.170 & 0.261 & [-784.067, 230.462] \\
    x6 & -6937.6588 & 1931.728 & -3.591 & 0.003 & [-1.11e+04, -2794.514] \\
    x7 & -0.0004 & 0.000 & -1.380 & 0.189 & [-0.001, 0.000] \\
    x8 & -0.0033 & 0.001 & -3.692 & 0.002 & [-0.005, -0.001] \\
    x9 & 0.0010 & 0.000 & 3.982 & 0.001 & [0.000, 0.001] \\
    \bottomrule
    \end{tabular}
    \caption{Absolute Regression Results: F1 Score}
    \label{tab:regression_absolute_f1}
    \end{table}
    
    \begin{itemize}
        \item Learning rate ($X_2$) had a significant positive impact on F1, with a coefficient of $0.0452$ ($p = 0.028$).
        \item Batch size ($X_3$) showed significance, with a coefficient of $0.1811$ ($p = 0.005$), reinforcing its role in F1 performance.
        \item The quadratic term for batch size ($X_8$) showed a significant negative effect ($p = 0.002$) on F1 in the absolute regression, indicating that excessively large batch sizes may begin to negatively impact F1 performance. This finding underscores the need to moderate batch size increases during the initial training phase to maintain optimal performance levels.
        \item The interaction between epochs and batch size ($X_9$) also displayed significance ($p = 0.001$), underlining the need for joint optimization.
    \end{itemize}

    \item \textbf{Relative Regression (Differences)}

    \begin{table}[H]
    \centering
    \scriptsize
    \renewcommand{\arraystretch}{1.1}
    \setlength{\tabcolsep}{4pt}
    \begin{tabular}{lrrrrr}
    \toprule
    Variable & Coef & Std Err & t & P>|t| & [0.025, 0.975] \\
    \midrule
    const & 0.7203 & 0.452 & 1.593 & 0.135 & [-0.257, 1.697] \\
    x1 & -1.181e+04 & 6232.671 & -1.895 & 0.080 & [-2.53e+04, 1650.901] \\
    x2 & 0.0250 & 0.015 & 1.684 & 0.116 & [-0.007, 0.057] \\
    x3 & 0.0244 & 0.017 & 1.450 & 0.171 & [-0.012, 0.061] \\
    x4 & 5.358e+08 & 3.07e+08 & 1.745 & 0.105 & [-1.28e+08, 1.2e+09] \\
    x5 & -382.7819 & 247.348 & -1.548 & 0.146 & [-917.144, 151.580] \\
    x6 & -6991.9007 & 1896.195 & -3.687 & 0.003 & [-1.11e+04, -2895.421] \\
    x7 & -0.0015 & 0.001 & -1.615 & 0.130 & [-0.003, 0.000] \\
    x8 & -0.0037 & 0.001 & -3.949 & 0.002 & [-0.006, -0.002] \\
    x9 & 0.0010 & 0.000 & 4.161 & 0.001 & [0.000, 0.001] \\
    \bottomrule
    \end{tabular}
    \caption{Relative Regression Results: F1 Score}
    \label{tab:regression_relative_f1}
    \end{table}
    
    \begin{itemize}
        \item The interaction between epochs and batch size ($X_9$) retained significance ($p = 0.001$), emphasizing its importance in fine-tuning.
        \item Learning rate ($X_2$) had a positive coefficient but with marginal significance ($p = 0.116$).
        \item The quadratic term for batch size (\( X_8 \)) demonstrated a significant negative impact on F1 (\( p = 0.002 \)), suggesting that larger batch sizes may start to adversely affect F1 performance when they exceed optimal values. This finding highlights the need to carefully balance batch size to avoid diminishing returns in F1 performance, especially as batch size increases.

    \end{itemize}
\end{itemize}

\paragraph{Visual Analysis and Insights}

The previous results are complemented by the Heatmaps (Figures 5 and 7) and Polynomial Regression Plots (Figures 6 and 8).

% \begin{itemize}
%     \item \textbf{Heatmap:} The heatmap (Figure \ref{fig:heatmap_absolute_f1}) shows correlations between F1 and the independent variables, with notable interactions across hyperparameters.

%     \item \textbf{Polynomial Regression Plot:} Figure \ref{fig:polynomial_regression_plot_absolute_f1} illustrates predicted vs. observed F1 scores, showing good model fit with a few outliers.

%     \item \textbf{Scatter Plots:} Scatter plots (Figure \ref{fig:scatter_plots_absolute_f1}) reveal clustering of F1 values across learning rates and batch sizes, emphasizing optimal configurations.
% \end{itemize}

\begin{figure*}[h!]
\centering
\begin{minipage}{0.40\textwidth}
    \centering
    \includegraphics[width=\textwidth]{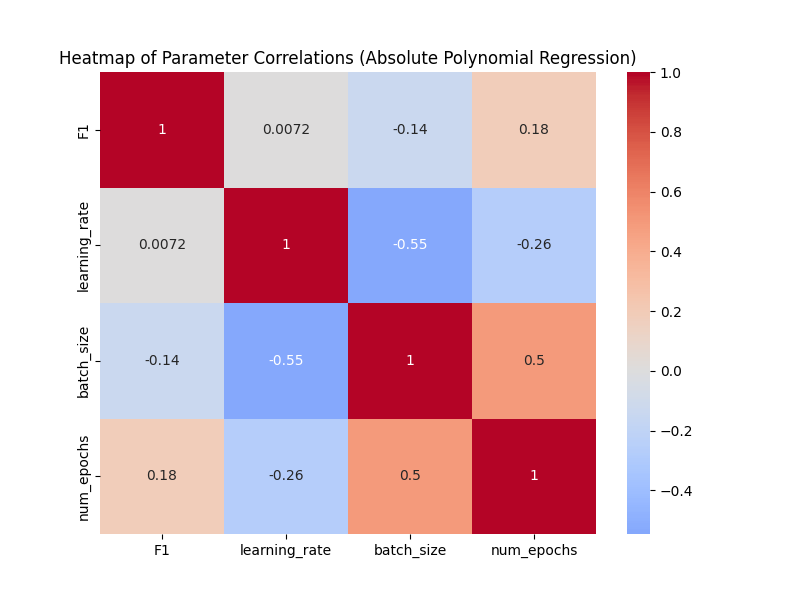}
    \caption{Heatmap of Parameter Correlations (Absolute Polynomial Regression - F1)}
    \label{fig:heatmap_absolute_f1}
\end{minipage}%
\hfill
%\end{figure*}
%\begin{figure*}[h]
%\centering
\begin{minipage}{0.40\textwidth}
    \centering
    \includegraphics[width=\textwidth]{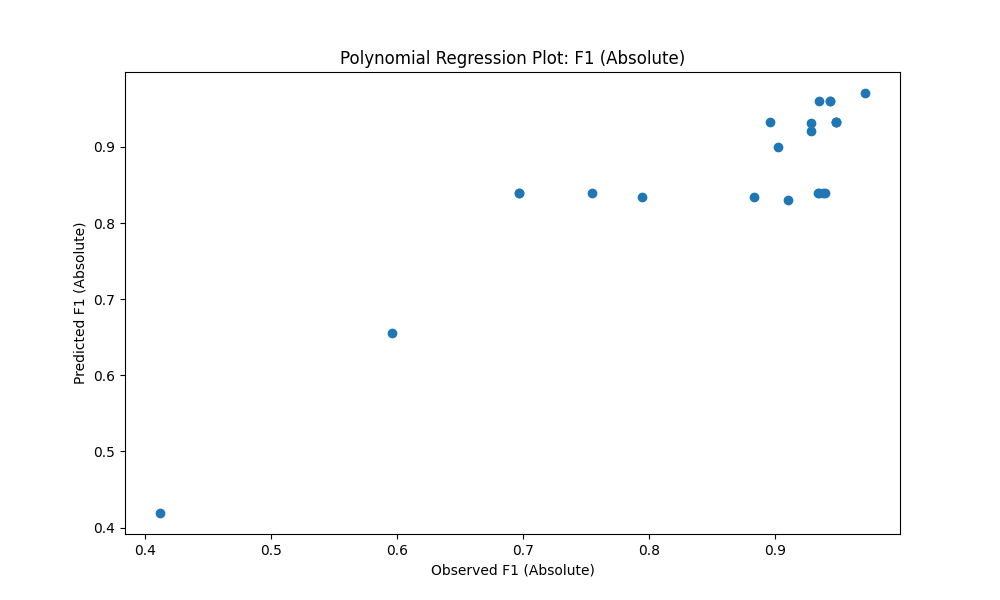}
    \caption{Polynomial Regression Plot: F1 Score (Absolute)}
    \label{fig:polynomial_regression_plot_absolute_f1}
\end{minipage}
\end{figure*}

%\vspace{0.5cm}

% \begin{figure*}[h!]
% \centering
% \begin{minipage}{0.40\textwidth}
%     \centering
%     \includegraphics[width=\textwidth]{figures/absolute_polynomial_scatter_plots_F1.png}
%     \caption{Scatter Plots: F1 vs Learning Rate, Batch Size, and Number of Epochs}
%     \label{fig:scatter_plots_absolute_f1}
% \end{minipage}%
%\hfill
%\end{figure*}
%\begin{figure*}[h]
%\centering
\begin{figure*}[h!]
\centering
\begin{minipage}{0.40\textwidth}
    \centering
    \includegraphics[width=\textwidth]{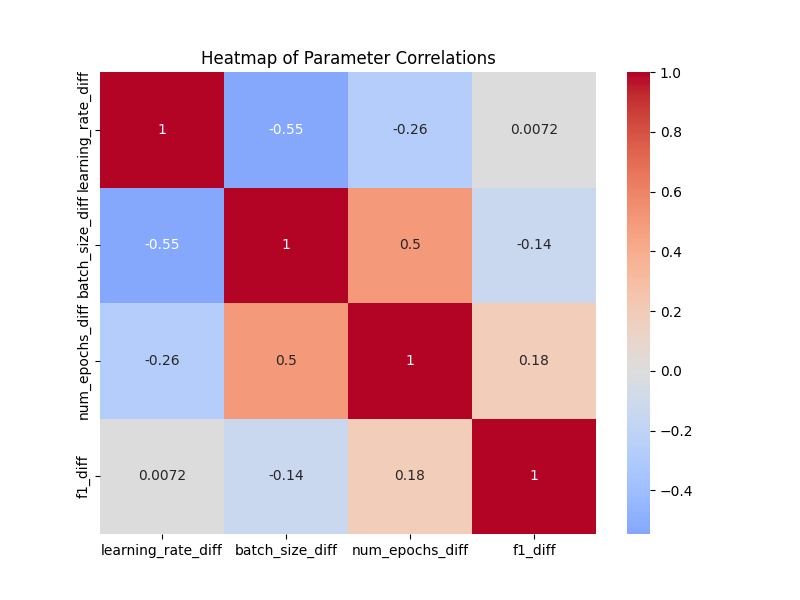}
    \caption{Heatmap of Parameter Correlations (Relative Polynomial Regression - F1)}
    \label{fig:heatmap_relative_f1}
\end{minipage}
\hfill
%\end{figure*}
%\includegraphics{figures/absolute_polynomial_heatmap_F1.png}
%\vspace{0.5cm}
%\begin{figure*}[h!]
%\centering
\begin{minipage}{0.40\textwidth}
    \centering
    \includegraphics[width=\textwidth]{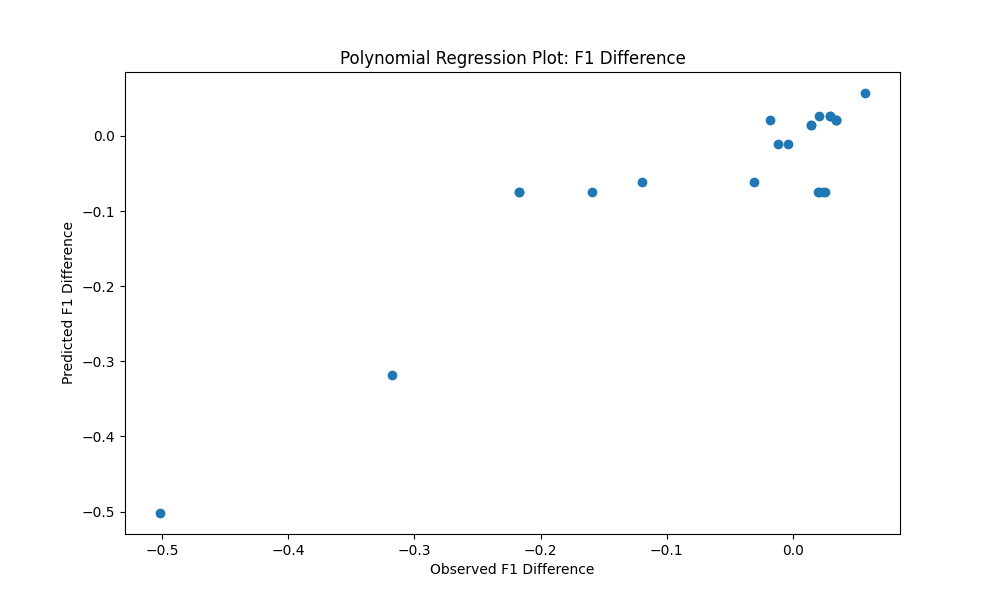}
    \caption{Polynomial Regression Plot: F1 Score Difference (Relative)}
    \label{fig:polynomial_regression_plot_relative_f1}
\end{minipage}%
%\hfill
\end{figure*}

% \begin{figure*}[h]
% \centering
% \begin{minipage}{0.40\textwidth}
%     \centering
%     \includegraphics[width=\textwidth]{figures/diff_polynomial_scatter_plots_F1.png}
%     \caption{Scatter Plots: F1 Score Difference vs Learning Rate, Batch Size, and Number of Epochs}
%     \label{fig:scatter_plots_relative_f1}
% \end{minipage}
% \end{figure*}

%\clearpage

\paragraph{Consolidated Findings and Practical Insights on F1-Score}

Our analysis identifies learning rate ($X_2$) and batch size ($X_3$) as key parameters for maximizing F1 performance, with significant positive impacts in the absolute analysis (p-values of 0.028 and 0.005, respectively), underscoring their importance in the initial training phase. However, the quadratic term for batch size ($X_8$) demonstrated a negative impact in both absolute and relative analyses (p = 0.002), indicating that while batch size improves F1 initially, excessive increases can reduce its efficacy, highlighting the need for careful batch size moderation. Additionally, the interaction between epochs and batch size ($X_9$) consistently showed significant positive effects in the relative analysis (p = 0.001), underscoring its role in enhancing F1 gains between fine-tuned and baseline models. Notably, learning rate ($X_2$) exhibited a consistent positive impact across both analyses, while batch size ($X_3$) showed higher significance in absolute terms (p = 0.005) but a diminished effect in relative analysis (p = 0.171), suggesting its greater influence on foundational rather than incremental performance.

These findings recommend a two-tiered approach: first, establishing a solid base model by fine-tuning learning rate and batch size to maximize initial F1; and second, moderating batch size increases due to diminishing returns, as indicated by the quadratic term for batch size. Fine-tuning should focus on the interaction between epochs and batch size to yield incremental F1 improvements, balancing foundational performance with efficient tuning for optimized outcomes.

\subsubsection{Loss – Comparison of Absolute and Relative Regressions}

\paragraph{Regression Results}

\begin{itemize}
    \item \textbf{Absolute Regression}

    \begin{table}[H]
    \centering
    \scriptsize
    \renewcommand{\arraystretch}{1.1}
    \setlength{\tabcolsep}{4pt}
    \begin{tabular}{lrrrrr}
    \toprule
    Variable & Coef & Std Err & t & P>|t| & [0.025, 0.975] \\
    \midrule
    x1 & -3776.1983 & 6.9e+04 & -0.055 & 0.957 & [-1.44e+05, 1.37e+05] \\
    x2 & -0.0368 & 0.071 & -0.515 & 0.610 & [-0.183, 0.109] \\
    x3 & 0.2311 & 0.164 & 1.406 & 0.170 & [-0.104, 0.566] \\
    x4 & 2.412e+08 & 1.13e+09 & 0.213 & 0.832 & [-2.06e+09, 2.55e+09] \\
    x5 & 1391.6662 & 1256.435 & 1.108 & 0.277 & [-1170.849, 3954.181] \\
    x6 & -3729.8604 & 5883.874 & -0.634 & 0.531 & [-1.57e+04, 8270.380] \\
    x7 & 0.0006 & 0.001 & 0.568 & 0.574 & [-0.002, 0.003] \\
    x8 & -0.0035 & 0.002 & -1.559 & 0.129 & [-0.008, 0.001] \\
    x9 & 0.0003 & 0.001 & 0.409 & 0.685 & [-0.001, 0.002] \\
    \bottomrule
    \end{tabular}
    \caption{Absolute Regression Results: Loss}
    \label{tab:regression_absolute_loss}
    \end{table}
    
    \begin{itemize}
        \item The learning rate ($X_2$) coefficient was -0.0368 with \(p = 0.610\), showing no statistical significance, indicating that variations in learning rate do not clearly impact absolute loss.
        \item Batch size ($X_3$) had a coefficient of 0.2311 with \(p = 0.170\), not statistically significant but suggesting a mild positive influence on loss.
        \item The quadratic term for num\_epochs ($X_4$) and batch size ($X_8$) had no practical relevance or significance.
    \end{itemize}

    \item \textbf{Relative Regression (Differences)}

    \begin{table}[H]
    \centering
    \scriptsize
    \renewcommand{\arraystretch}{1.1}
    \setlength{\tabcolsep}{4pt}
    \begin{tabular}{lrrrrr}
    \toprule
    Variable & Coef & Std Err & t & P>|t| & [0.025, 0.975] \\
    \midrule
    const & -4.6383 & 1.891 & -2.453 & 0.021 & [-8.511, -0.766] \\
    x1 & 1.735e+04 & 3.4e+04 & 0.511 & 0.614 & [-5.22e+04, 8.69e+04] \\
    x2 & -0.1531 & 0.066 & -2.327 & 0.027 & [-0.288, -0.018] \\
    x3 & -0.0105 & 0.074 & -0.142 & 0.888 & [-0.163, 0.141] \\
    x4 & -1.349e+09 & 1.25e+09 & -1.082 & 0.289 & [-3.9e+09, 1.21e+09] \\
    x5 & 2843.7941 & 1335.126 & 2.130 & 0.042 & [108.913, 5578.675] \\
    x6 & 6416.8155 & 6955.611 & 0.923 & 0.364 & [-7831.109, 2.07e+04] \\
    x7 & 0.0096 & 0.004 & 2.517 & 0.018 & [0.002, 0.017] \\
    x8 & 0.0041 & 0.004 & 1.092 & 0.284 & [-0.004, 0.012] \\
    x9 & -0.0013 & 0.001 & -1.410 & 0.170 & [-0.003, 0.001] \\
    \bottomrule
    \end{tabular}
    \caption{Relative Regression Results: Loss}
    \label{tab:regression_relative_loss}
    \end{table}
    
    \begin{itemize}
        \item The learning rate ($X_2$) had a significant coefficient of -0.1531 with \(p = 0.027\), indicating that increases in learning rate can reduce the loss difference between fine-tuned and base models.
        \item Parameter $X_5$ (interaction between learning rate and number of epochs) was statistically significant with a coefficient of 2843.79 and \(p = 0.042\). This suggests that specific optimizations involving learning rate and epochs have a notable impact on reducing the loss difference between fine-tuned and baseline models.

        % \item Parameter $X_5$ (nteraction between learning rate and number of epochs) was significant with a coefficient of 2843.79 and \(p = 0.042\), suggesting specific optimizations impact the loss difference.
        \item The quadratic term for learning rate ($X_7$) had a significant impact with \(p = 0.018\), highlighting the importance of non-linear adjustments in learning rate on loss differences.
    \end{itemize}
\end{itemize}

\paragraph{Visual Analysis and Insights}

The previous results are complemented by the Heatmaps (Figures 9 and 11) and Polynomial Regression Plots (Figures 10 and 12). 

% \begin{itemize}
%     \item \textbf{Heatmap:} The heatmap (Figure \ref{fig:heatmap_absolute_loss}) shows moderate negative correlations between learning rate and batch size (-0.46) and weak correlations between num\_epochs and loss.
    
%     \item \textbf{Polynomial Regression Plot:} Figure \ref{fig:polynomial_regression_plot_absolute_loss} shows significant dispersion, especially for higher loss values, indicating that the model may not fully capture variability at the extremes.

%     \item \textbf{Scatter Plots:} Scatter plots (Figure \ref{fig:scatter_plots_absolute_loss}) highlight variability in loss, especially for lower values of learning rate and batch size, indicating sensitivity to these parameters.
% \end{itemize}

\begin{figure*}[h!]
\centering
\begin{minipage}{0.40\textwidth}
    \centering
    \includegraphics[width=\textwidth]{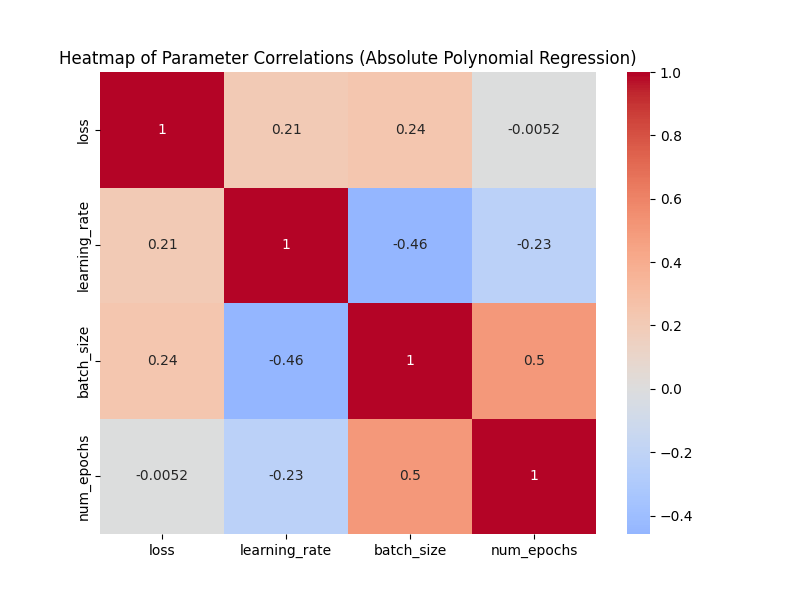}
    \caption{Heatmap of Parameter Correlations (Absolute Polynomial Regression)}
    \label{fig:heatmap_absolute_loss}
\end{minipage}%
\hfill
%\end{figure*}
%\begin{figure*}[h]
%\centering
\begin{minipage}{0.40\textwidth}
    \centering
    \includegraphics[width=\textwidth]{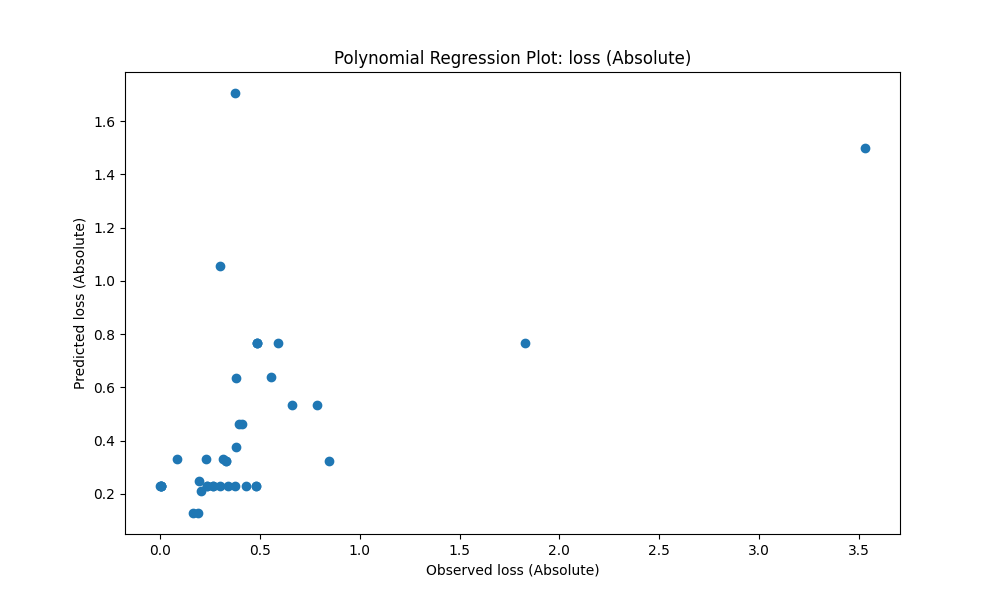}
    \caption{Polynomial Regression Plot: Loss (Absolute)}
    \label{fig:polynomial_regression_plot_absolute_loss}
\end{minipage}
\end{figure*}

%\vspace{0.5cm}
% \begin{figure*}[h!]
% \centering
% \begin{minipage}{0.40\textwidth}
%     \centering
%     \includegraphics[width=\textwidth]{figures/absolute_polynomial_scatter_plots_LOSS.png}
%     \caption{Scatter Plots: Loss vs Learning Rate, Batch Size, and Number of Epochs}
%     \label{fig:scatter_plots_absolute_loss}
%\end{minipage}%
%\hfill
%\end{figure*}
%\begin{figure*}[h]
%\centering
\begin{figure*}[h!]
\centering
\begin{minipage}{0.40\textwidth}
    \centering
    \includegraphics[width=\textwidth]{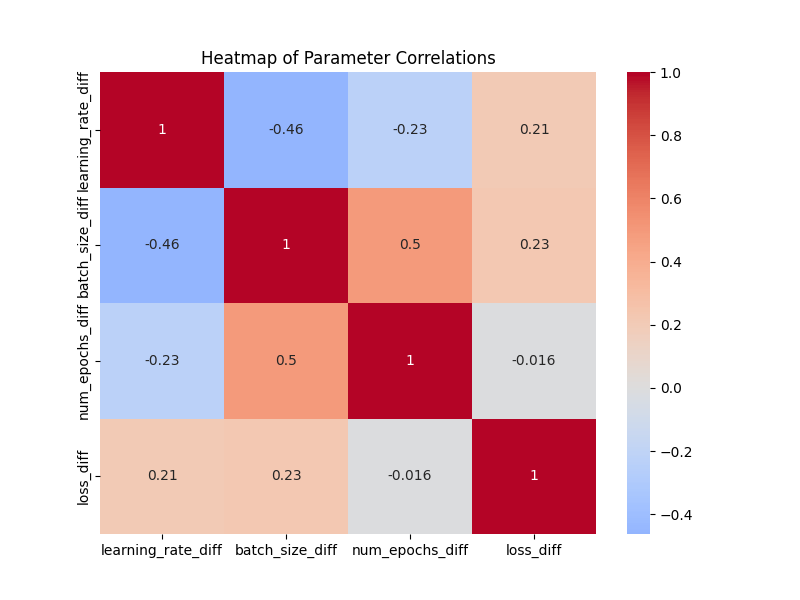}
    \caption{Heatmap of Parameter Correlations (Relative Polynomial Regression)}
    \label{fig:heatmap_relative_loss}
\end{minipage}
%\end{figure*}
%\vspace{0.5cm}
\hfill
%\begin{figure*}[h!]
%\centering
\begin{minipage}{0.40\textwidth}
    \centering
    \includegraphics[width=\textwidth]{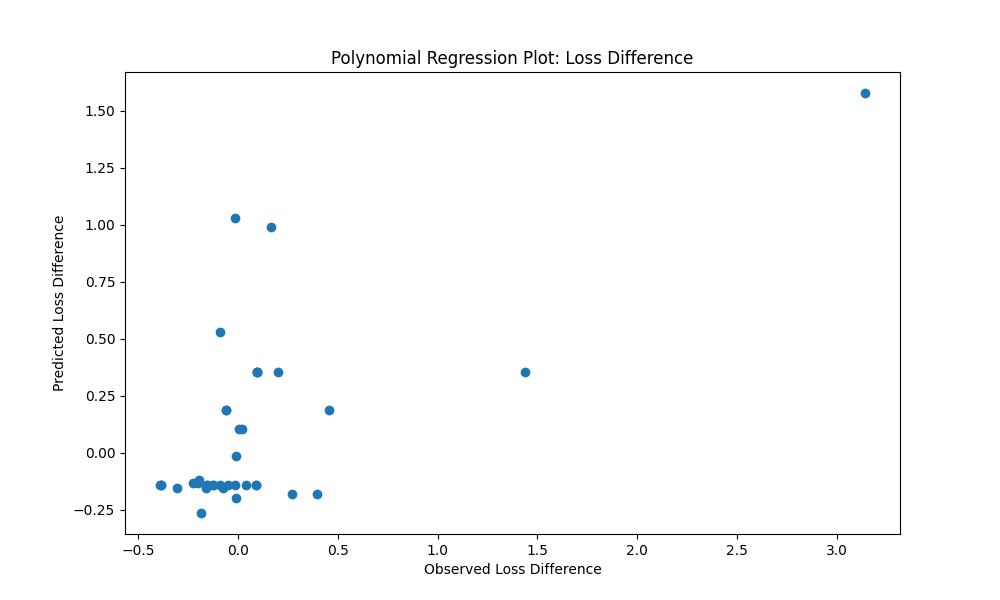}
    \caption{Polynomial Regression Plot: Loss Difference (Relative)}
    \label{fig:polynomial_regression_plot_relative_loss}
\end{minipage}
%\hfill
\end{figure*}

% \begin{figure*}[h]
% \centering
% \begin{minipage}{0.40\textwidth}
%     \centering
%     \includegraphics[width=\textwidth]{figures/diff_polynomial_scatter_plots_LOSS.png}
%     \caption{Scatter Plots: Loss Difference vs Learning Rate, Batch Size, and Number of Epochs}
%     \label{fig:scatter_plots_relative_loss}
% \end{minipage}
% \end{figure*}

%\clearpage

\paragraph{Findings and Practical Insights on Loss}

Our analysis of loss indicates that, in the absolute context, no single parameter had a significant impact, suggesting that broader hyperparameter adjustments alone may not account for loss variability. However, in the relative analysis, learning rate ($X_2$) showed a significant negative impact ($p = 0.027$), implying that moderate increases in learning rate can help reduce loss. The quadratic term for learning rate ($X_7$) highlights the relevance of non-linear adjustments for fine-tuning. Furthermore, the interaction between learning rate and epochs ($X_5$) exhibited a statistically significant positive effect (coefficient = 2843.79, $p = 0.042$), suggesting that coordinated adjustments of these parameters can meaningfully decrease incremental loss, underscoring the need for a balanced fine-tuning strategy.

Based on these findings, a strategic approach is recommended: initially focusing on batch size and epochs to stabilize training and control loss variability, followed by precise adjustments to learning rate and its interactions to optimize incremental performance during fine-tuning. Coordinating learning rate and epochs allows for more controlled and effective loss reduction, particularly valuable in settings where nuanced improvements are desired.

\subsection{Mapping Fine-Tuning Strategies and Recommendations for Improvements}

% \subsubsection{Objective}
% This section integrates practical insights and conclusions from the analyses in the previous subsections on accuracy, F1, and loss, evaluating cross-effects of optimization strategies. This integration includes mapping consistencies and inconsistencies, along with practical recommendations for efficient and balanced adjustments across metrics.

\subsubsection{Strategy Mapping by Metric}
The following are suggested strategies for optimizing each metric and their potential cross-impacts on the others.

\paragraph{1. Strategies Focused on Accuracy}
\begin{itemize}
    \item \textbf{Key Parameters:}
    \begin{itemize}
        \item \textbf{Batch Size (X2):} Consistent and significant impact (\(p = 0.028\)) in absolute analysis, crucial for establishing a solid performance foundation.
        \item \textbf{Num\_epochs (X3):} Essential for overall accuracy performance (\(p = 0.005\)).
        \item \textbf{Learning Rate (X7):} Relevant in the relative analysis, with a critical role in fine adjustments (\(p = 0.004\)).
    \end{itemize}
    \item \textbf{Cross-Effects:}
    \begin{itemize}
        \item \textbf{F1:} Prioritizing batch size and epochs tends to improve F1, but optimizing learning rate will be necessary to capture incremental gains.
        \item \textbf{Loss:} While these strategies may contribute to consistent training, there is a risk of increased loss variability without adequate regularization techniques.
    \end{itemize}
\end{itemize}

\paragraph{2. Strategies Focused on F1 Score}
\begin{itemize}
    \item \textbf{Key Parameters:}
    \begin{itemize}
        \item \textbf{Learning Rate (X2):} Significant positive impact in absolute analysis (\(p = 0.028\)).
        \item \textbf{Batch Size (X3):} Crucial for F1 performance (\(p = 0.005\)), though with reduced impact in relative analysis.
        \item \textbf{Interaction between Epochs and Batch Size (X9):} Essential for maximizing F1 (\(p = 0.001\)).
    \end{itemize}
    \item \textbf{Cross-Effects:}
    \begin{itemize}
        \item \textbf{Accuracy:} Joint optimization of epochs and batch size can benefit accuracy.
        \item \textbf{Loss:} Fine adjustments can improve overall performance, but there is a risk of overfitting, increasing loss.
    \end{itemize}
\end{itemize}

\paragraph{3. Strategies Focused on Reducing Loss}
\begin{itemize}
    \item \textbf{Key Parameters:}
    \begin{itemize}
        \item \textbf{Learning Rate (X2):} Significantly reduces loss in relative analysis (\(p = 0.027\)).
        \item \textbf{Quadratic Term of Learning Rate (X7):} Relevant impact in fine adjustments (\(p = 0.018\)).
        \item \textbf{Batch Size (X3):} Relevant but not significant.
    \end{itemize}
    \item \textbf{Cross-Effects:}
    \begin{itemize}
        \item \textbf{Accuracy and F1:} Reducing loss may not directly optimize these metrics, requiring balance to avoid compromising overall performance.
    \end{itemize}
\end{itemize}

\subsubsection{Identified Consistencies and Inconsistencies}

\paragraph{Batch Size (X2)}
\begin{itemize}
    \item \textbf{Consistency:} Positively impacts both accuracy and F1 in absolute analysis.
    \item \textbf{Inconsistency:} While relevant, it was not significant for loss, suggesting limited effect on this metric.
\end{itemize}

\paragraph{Learning Rate (X7)}
\begin{itemize}
    \item \textbf{Consistency:} Fine adjustments are critical for incremental optimization in accuracy and loss.
    \item \textbf{Inconsistency:} In the relative analysis of F1, it showed marginal significance, indicating a lesser incremental impact.
\end{itemize}

\paragraph{Interaction between Epochs and Batch Size (X9)}
\begin{itemize}
    \item \textbf{Consistency:} Essential for optimizing F1.
    \item \textbf{Inconsistency:} Had limited impact on accuracy and loss.
\end{itemize}

\subsubsection{Practical Recommendations and Fine-Tuning Strategies}

Based on the previous analyses, we propose an incremental and integrated optimization approach:

\paragraph{Initial Selection}
\begin{itemize}
    \item \textbf{Batch Size and Epochs:} Prioritize these hyperparameters in the initial phase to ensure a solid foundation, improving accuracy and F1.
\end{itemize}

\paragraph{Final Adjustments}
\begin{itemize}
    \item \textbf{Learning Rate:} Carefully refine to capture incremental gains in loss and incremental performance in F1.
\end{itemize}

\paragraph{Trade-Offs and Cross-Impact}
\begin{itemize}
    \item \textbf{Joint Monitoring:} When optimizing one metric, it is essential to monitor the others. For example, a high learning rate may reduce loss but compromise accuracy.
    \item \textbf{Regularization:} Apply regularization techniques to prevent overfitting and ensure that reducing loss does not harm overall performance.
\end{itemize}

\paragraph{Interactions and Non-Linearities:}
\begin{itemize}
    \item \textbf{Explore Interactions:} Adjust epochs and batch size jointly to optimize F1.
    \item \textbf{Normalization and Regularization:} Reduce variability at extremes and enhance model robustness.
\end{itemize}

\section{Conclusions}
This study presents an in-depth evaluation of fine-tuning strategies for the text classification task, focusing on the DistilBERT model variant distilbert-base-uncased-finetuned-sst-2-english. By conducting polynomial regression analyses on three primary evaluation metrics—accuracy, F1-score, and loss—and examining the roles of key hyperparameters (learning rate, batch size, and number of epochs), our findings emphasize the complex interplay and trade-offs inherent in hyperparameter tuning for large language models (LLMs).

The analysis revealed that optimizing one metric often introduces variability or trade-offs that affect other metrics, particularly when pursuing fine-tuning for incremental improvements. For example, while batch size consistently enhanced both accuracy and F1-score in absolute analysis, its impact on loss optimization was limited. In contrast, learning rate adjustments, especially in relative analysis, showed significant reductions in loss (p = 0.027) but required careful balancing to avoid performance inconsistencies in accuracy. The consistent significance of the interaction between epochs and batch size for F1-score maximization (p = 0.001) further emphasizes the necessity of joint optimization to maintain balanced performance across metrics.

Beyond these core interdependencies, this study underscores the importance of understanding variability in performance metrics across different contexts and tasks beyond text classification. The non-linear and, at times, inconsistent effects of hyperparameters observed here suggest that fine-tuning strategies for LLMs should be adaptive, potentially varying by both metric and the unique demands of each subtask or broader task domain, such as NLP or computer vision. This adaptability is essential for maximizing model robustness and efficacy across diverse deployment scenarios.

Our findings also highlight the potential of developing fine-tuning frameworks that incorporate metric variability and dynamically adjust hyperparameters based on the model's performance across multiple metrics. Such frameworks, grounded in empirical data on hyperparameter impacts, provide a promising foundation for the sustainable and optimized development of LLMs in a wide array of applications.

 \end{document}